\def\year{2019}\relax
\documentclass[letterpaper]{article} %DO NOT CHANGE THIS
\usepackage{aaai19}  %Required
\usepackage{times}  %Required
\usepackage{helvet}  %Required
\usepackage{courier}  %Required
\usepackage{url}  %Required
\usepackage{graphicx}  %Required

\usepackage{algorithm}
\usepackage{algpseudocode}
\usepackage[T1]{fontenc}
\usepackage[utf8]{inputenc}
\usepackage{amsmath}
\usepackage{amssymb}
\usepackage{wasysym}
\usepackage{amsfonts}
\usepackage{array}
\usepackage{multirow}
\usepackage{caption}
\usepackage{color}
\usepackage[inline]{enumitem}

\newcommand{\smallsize}{\fontsize{8.3pt}{7.3pt}\selectfont}

\begin{document}
% The file aaai.sty is the style file for AAAI Press 
% proceedings, working notes, and technical reports.
%
\title{Neural Task Representations as Weak Supervision for Model Agnostic Cross-Lingual Transfer}
\author{Sujay Kumar Jauhar \\
Microsoft Research AI \\
Redmond, WA, USA \\
{\tt sjauhar@microsoft.com} \\\And
Michael Gamon \\
Microsoft Research AI \\
Redmond, WA, USA \\
{\tt mgamon@microsoft.com} \\\And
Patrick Pantel\thanks{This work was done while the third author was at Microsoft Research AI.} \\
Facebook Inc. \\
Seattle, WA, USA \\
{\tt ppantel@fb.com}}

\maketitle
\begin{abstract}
Natural language processing is heavily Anglo-centric, while the demand for models that work in languages other than English is greater than ever. Yet, the task of transferring a model from one language to another can be expensive in terms of annotation costs, engineering time and effort. In this paper, we present a general framework for easily and effectively transferring neural models from English to other languages. The framework, which relies on task representations as a form of weak supervision, is model and task agnostic, meaning that many existing neural architectures can be ported to other languages with minimal effort. The only requirement is unlabeled parallel data, and a loss defined over task representations. We evaluate our framework by transferring an English sentiment classifier to three different languages. On a battery of tests, we show that our models outperform a number of strong baselines and rival state-of-the-art results, which rely on more complex approaches and significantly more resources and data. Additionally, we find that the framework proposed in this paper is able to capture semantically rich and meaningful representations across languages, despite the lack of direct supervision.
\end{abstract}

\section{Introduction}\label{sec:intro}

Recent advances in Natural Language Processing (NLP) and Deep Learning have enabled systems to achieve human parity in several key research areas such as Speech Recognition~\cite{xiong2016achieving}, Machine Translation~\cite{hassan2018achieving} and Machine Comprehension~\cite{hu2017mnemonic}, while advancing the state-of-the-art for many other tasks. Nevertheless, much of this research revolves around models, methods and datasets that are Anglo-centric. It is estimated\footnote{\url{https://en.wikipedia.org/wiki/English-speaking_world}} that only about 350 million people are native English speakers, while another 500 million to 1 billion speak it as a second language. This accounts for at most 20\% of the world's population. With language technologies making inroads into the digital lives of people, it is becoming increasingly imperative to build NLP applications that can understand the other 80\% of the world.

Building such systems from scratch can be expensive and time-consuming. Performant NLP models often rely on
vast amounts of high quality annotated data, which come at the cost of annotator time, effort and money. Therefore, much of the research community's efforts towards building tools for other languages have relied on transferring existing English models to other languages.

Previous efforts have relied on Machine Translation (MT) to translate training or test data from English to a target language~\cite{wan2009co,shi2010cross}. Others have additionally considered bilingual dictionaries to directly transfer features~\cite{mihalcea2007learning}, while some have explored methods such as structural correspondence learning~\cite{prettenhofer2010cross}, and cross-lingual representation learning~\cite{huang2013cross}.

Building state-of-the-art MT systems requires expertise and vast amounts of training data. The best commercial APIs such as Google Translate\footnote{\url{https://cloud.google.com/translate/}} and Microsoft Translator\footnote{\url{https://www.microsoft.com/en-us/translator/translatorapi.aspx}} charge on a per-character basis, and the long-term costs for model transfer can become prohibitive. Meanwhile, bilingual dictionaries can be equally expensive to build if done manually, or contain significant noise, if induced automatically.

In this paper we propose a novel framework to transfer an existing neural English model to another language with minimal cost and effort. The framework leverages \emph{task representations} -- the layer in a neural network model immediately preceding the prediction layer, which we hypothesize captures an abstract semantic representation of the prediciton problem -- as a form of weak supervision in unlabeled data. The result is a model that learns a language agnostic neural architecture, jointly with language specific embeddings.

Specifically, the framework:
\begin{enumerate*}[label=(\roman*)]
    \item Is model and task agnostic, and thus applicable to a wide range of existing neural architectures.
    \item Needs no target language training data, no translation system or bilingual dictionary -- only a parallel corpus.
    \item And has the sole modelling requirement of defining a loss over task representations, thereby greatly reducing the engineering effort involved in model transfer. 
\end{enumerate*}

Our approach is especially useful when a high quality MT system is not available for a target language or specialized domain. However, we also show in our experiments that it can be used in conjunction with an MT based transfer approach, to produce state-of-the-art results.  

We test our framework on transferring an English sentiment classifier, and evaluate on 3 different target languages using a public dataset. Our experiments show that our approach not only outperforms several strong baselines (including an MT system trained on the same parallel data) and previous approaches, but even outperforms the previous state-of-the-art on one language, while relying on a much simpler method and requiring significantly fewer resources. Additionally, our approach is able to learn semantically meaningful cross-lingual representations without the need for any direct supervision.
\section{Related Work}\label{sec:related}

There are several research areas related to work in this paper. Cross-lingual model transfer can be more generally perceived as an instance of domain adaptation. Previous work in this area has explored techniques such as structural correspondence learning~\cite{blitzer2006domain}, and feature sharing via augmentation~\cite{daume2009frustratingly}. Domain adaptation for deep learning has also been tackled, for example with stacked de-noising autoencoders~\cite{glorot2011domain}, a gradient reversal layer for domain differentiation~\cite{ganin2014unsupervised} and generalizations of Convolutional Neural Networks that capture consistencies between domains~\cite{long2015learning}.

\cite{yosinski2014transferable} study the transferability of neural network components in the context of image recognition. They suggest that higher layers tend to be more specialized and domain specific, and therefore less generalizable.~\cite{mou2016transferable} investigate the same problem, but looking more specifically at domain adaptation in NLP. They find that all but the embedding layer tend to be too domain specific to be practically transferable. Nevertheless what we propose in our framework for cross-lingual transfer is the opposite: specifically, sharing the higher layers of a network while maintaining language specific embeddings. Our results demonstrate that this approach is, in fact, successful for our evaluation task of sentiment classification.

Sharing information across domains also pertains to Multi-Task Learning. The deep learning community's work in this area can be broadly separated into two approaches: hard, and soft parameter sharing. In hard parameter sharing, models share a common architecture with some task-specific layers~\cite{zhang2014facial,long2015learning2}, while in soft parameter sharing tasks have their own sets of parameters that are constrained by some shared penalty~\cite{yang2016trace,misra2016cross}. Our work fits in the first category, with a shared model that is complemented by language specific embeddings.

Several efforts have focused specifically on cross-lingual model transfer for NLP tasks such as part-of-speech tagging~\cite{das2011unsupervised}, speech recognition~\cite{huang2013cross} and semantic role labeling~\cite{pado2009cross}. A number of researchers have also looked at sentiment classification, such as~\cite{wan2009co},~\cite{pan2011cross},~\cite{xiao2014semi} and~\cite{zhou2016attention}. While we evaluate our approach of model transfer on sentiment classification, our paper proposes a general framework capable of tackling many different NLP problems in a unified way.

Different techniques have been used for cross-lingual transfer, including co-training~\cite{wan2009co}, structural correspondence learning~\cite{prettenhofer2010cross}, and matrix factorization~\cite{pan2011cross,xiao2014semi}. More recently, neural approaches have also tackled the problem of model transfer with approaches such as layer sharing across languages~\cite{huang2013cross}, and adversarial training~\cite{chen2016adversarial}.

The closest to our work, in terms of modeling paradigm, is prior research on label projection~\cite{mihalcea2007learning,das2011unsupervised}, feature projection~\cite{kozhevnikov2014cross} and weak supervision~\cite{shi2010cross,lee2013pseudo}. Our work differs from these methods by being a neural framework that integrates task featurization, model learning and cross-lingual transfer in a joint schema, while being flexible enough to accommodate a wide range of target applications.

We also note research on bilingual embeddings, which relates to model transfer by seeking to embed units from different languages in a joint space. Previous solutions include modeling with canonical correlation analysis~\cite{faruqui2014improving}, matrix factorization~\cite{shi2015learning} and neural networks~\cite{klementiev2012inducing}. The popular Word2Vec model~\cite{mikolov2013efficient}, for example, has been modified to work for bilingual word embeddings~\cite{gouws2015bilbowa,luong2015bilingual}, while others have tackled longer units of representations~\cite{pham2015learning}.~\cite{upadhyay2016cross} provides a comparison of several of these approaches.
Notably, these efforts typically only deal with the first embedding layer of neural networks, rather than full model transfer -- which is the problem tackled in this paper.

\begin{figure*}[ht]
\centering
\includegraphics[width=0.99\textwidth]{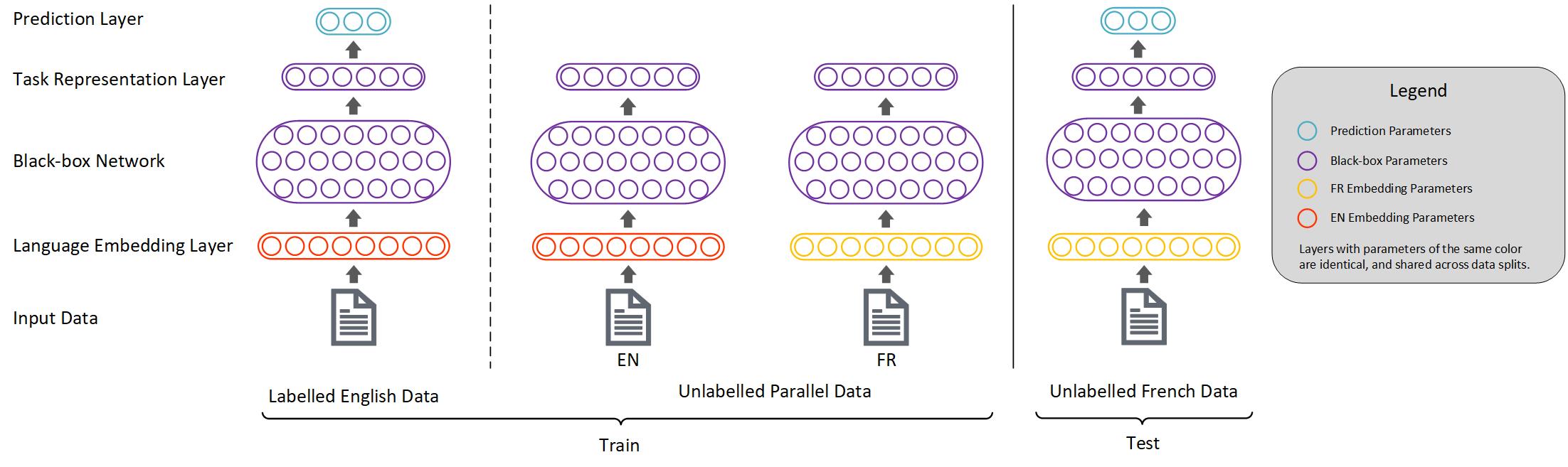}
\caption{The setup of the data and model depicting our general framework for transferring an English model to French. English embeddings are red, while French embeddings are gold. The shared black-box architecture is purple, and the predictive output layer is blue.}
\label{fig:architecture}
\end{figure*}
\section{Model Transfer by Representation Projection}\label{sec:model}

To describe our general framework for cross-lingual neural model transfer we begin by providing the intuition behind the framework, then formalizing it. Finally we show how it can be applied to easily transfer an English sentiment classification model to other languages.

\subsection{General Framework}\label{subec:frmwrk}

Many neural NLP models can be decomposed into three distinct components. The first is an embedding layer, which converts linguistic units (words, characters, etc.) into dense representations. The last is a prediction layer, -- often a softmax function -- used to produce a probability distribution over the space of output labels. Sandwiched in-between these two is some task-appropriate model architecture. Our general framework is agnostic to the choice of this architecture, which can have an arbitrary network configuration and number of layers. Therefore, let us assume, for the sake of this exposition, that it is a black-box.
In particular, we draw attention to the last of these layers (i.e. the one immediately preceding the prediction layer) and call this the \emph{task representation layer}. These components are depicted in Figure~\ref{fig:architecture}.

Our approach for model transfer relies on two key hypotheses. First, the black-box network and prediction layer can be shared across languages; this means that the only difference between an English and French model (for example) is the language specific embeddings (shown by color in Figure~\ref{fig:architecture}). Second, the task representation layer captures all the information required to make a successful prediction.

Now consider a model trained on English labeled examples, and some parallel data. An indication of successful model transfer is that the French model predicts the same \emph{thing} on the French side of parallel data as the English model does on the English side. We do not really care what this \emph{thing} is. It can be an actual label~\cite{mihalcea2007learning}, in which case the approach can be referred to as \emph{label projection}. But we can also try to produce the same task representation in both languages (\emph{representation projection}), a softer form of weak supervision. This is the core of our framework. We will show in our experiments (see Section~\ref{sec:eval}) that soft representation projection is significantly better than label projection.

We now formalize these intuitions. Consider a task $\mathbb{T}$ with some labeled data $D_L = \{(x_i, y_i) \mid 0 \leq i \leq N\}$, where $x_i$ are English inputs and $y_i$ are outputs that take on $K$ possible values. Without loss of generality let us assume that the inputs $x_i = \{e_{i1},...,e_{il}\}$ are sequences of English words\footnote{Our framework is agnostic to the granularity of units, and is applicable regardless of this choice.}. Moreover we have some parallel data $D_P = \{(e_j, f_j) \mid 0 \leq j \leq M\}$, where $e_j = \{e_{j1},...,e_{jl_1}\}$ and $f_j = \{f_{j1},...,f_{jl_2}\}$ are parallel English and French sentences respectively.

Let us define English embeddings $U = \{\vec{u_i} \mid \forall e_i \in V_E\}$ such that there is a vector for every word in vocabulary $V_E$. Similarly let us define $V = \{\vec{v_i} \mid \forall f_i \in V_F\}$ for vocabulary $V_F$. To enable a shared model architecture, we require that dimensions $d$ of vectors $\vec{u_i}$ and $\vec{v_i}$ be the same. We denote a mapping of the English sequence $e_j = \{e_{j1},...,e_{jm}\}$ to a sequence of vectors as $\vec{\vec{u_j}}$ and a French sequence $f_j = \{f_{j1},...,f_{jn}\}$ to a sequence of vectors $\vec{\vec{v_j}}$. We define a black-box model $\mathcal{M}$ with parameters $\theta_\mathcal{M}$ that takes as input, a sequence of embeddings, and yields a task representation. Specifically, for an English input $x_i$:

\begin{equation}\label{eq:bbmodel}
\mathcal{R}_{x_i}^\mathbb{T} = \mathcal{M}(\vec{\vec{u_i}};\theta_\mathcal{M})
\end{equation}

\noindent Finally we consider a prediction layer $\mathcal{\pi}$ with parameters $\theta_\mathcal{\pi}$ that yields a probability distribution over the $K$ output variables:

\begin{equation}\label{eq:pred}
\widehat{\mathcal{\pi}_i^k} = \frac{\mathcal{\pi}_k(\mathcal{R}_{x_i}^\mathbb{T};\theta_\mathcal{\pi})}{\displaystyle  \sum_{j = 1}^K \mathcal{\pi}_j(\mathcal{R}_{x_i}^\mathbb{T};\theta_\mathcal{\pi})}
\end{equation}

\noindent where $\pi_k$ is the $k^\text{th}$ neuron of the layer, and we use the shorthand $\widehat{\mathcal{\pi}_i^k}$ to denote $P(\hat{y_i} = k)$. Our framework then optimizes two losses.

\textbf{Labeled Loss:} Given that we have labeled English data, we optimize the following loss for the combined network:

\begin{equation}\label{eq:labloss}
\mathcal{L}_{D_L} = \displaystyle \sum_{i = 1}^N \sum_{k = 1}^K \Delta_L \left(\widehat{\mathcal{\pi}_i^k}, y_i\right)
\end{equation}

\noindent where $\Delta_L$ is a loss function defined between $\widehat{\mathcal{\pi}_i^k}$ and the response variable $y_i$. For example, in the binary case $\Delta_L$ might be a cross-entropy loss.

\textbf{Unlabeled Loss}: On the parallel data we use the English task representations generated by the model as weak supervision for the French side. Specifically:

\begin{equation}\label{eq:parloss}
\mathcal{L}_{D_P} = \displaystyle \sum_{j = 1}^M \Delta_P \left(\mathcal{R}_{e_j}^\mathbb{T}, \mathcal{R}_{f_j}^\mathbb{T} \right)
\end{equation}

\noindent where $\Delta_P$ is some loss function between task representations yielded on parallel inputs. Since task representations are vectors, the mean-squared error between them -- for example -- might be an appropriate loss.

Then jointly, our final optimization problem is $\mathcal{L} = \mathcal{L}_{D_L} + \alpha \mathcal{L}_{D_P}$, where $\alpha$ is a hyperparameter that controls the mixing strength between the two loss components.

Notice that with this framework there is no requirement for MT, since neither training nor test data is ever translated. Nor are any other resources, such as a pivot lexicon or bilingual dictionary, used. The only requirement is paralled data and the definition of a loss function $\mathcal{L}_{D_P}$. The model architecture $mathcal{M}$ and the labeled loss $\mathcal{L}_{D_L}$ are properties one would have defined anyway for an English only model.

\begin{figure}
\centering
\includegraphics[width=0.49\textwidth]{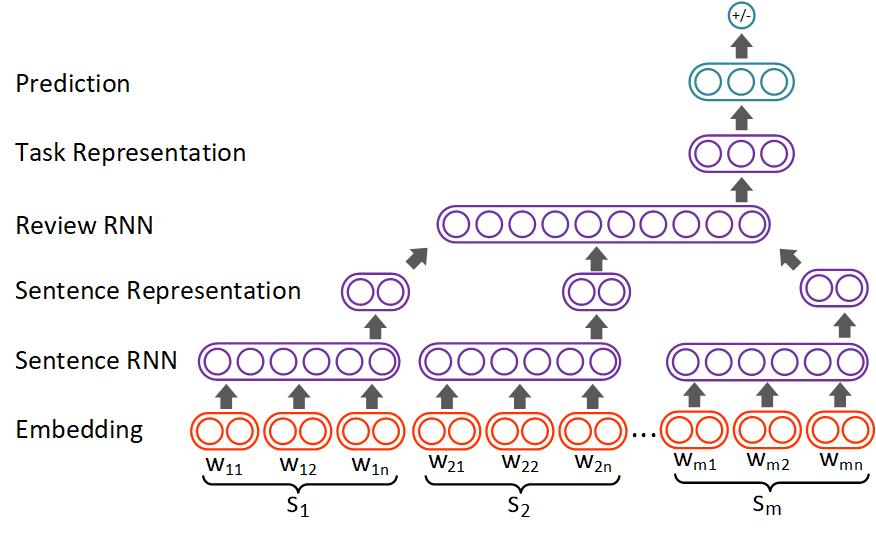}
\caption{The hierarchical RNN model architecture that we use to predict sentiment polarity. A sentence RNN is used to convert sequences of word embeddings into sentence representations, which are in turn input into a %\mg{"Review-level" maybe?} SKJ: fixed.
Review-level RNN to generate a task representation.}
\label{fig:sentirnn}
\end{figure}

\subsection{Training and Prediction}\label{subsec:training}

With well defined loss functions $\Delta_L$ and $\Delta_P$, training consists of back-propagating errors through the network and updating the parameters of the model. We distinguish two separate scenarios: two-stage training, and joint training.

In two-stage training we first train the labeled loss given by Equation~\ref{eq:labloss} on the labeled English data by finding $U^*, \theta_\mathcal{M}^*, \theta_\pi^* = \displaystyle \arg\max_{U, \theta_\mathcal{M}, \theta_\pi} \mathcal{L}_{D_L}$. We then freeze the English embeddings $U$, and the shared model parameters $\theta_\mathcal{M}$ and $\theta_\pi$. In the second stage we only update the French embeddings $V$ on the parallel data using the unlabeled projection loss given by Equation~\ref{eq:parloss} by optimizing $V^* = \displaystyle \arg\max_{V} \mathcal{L}_{D_P}$.

In contrast, for joint training, we do not freeze any parameters and make joint updates. That is, we optimize $U^*, V^*, \theta_\mathcal{M}^*, \theta_\pi^* = \displaystyle \arg\max_{U, V \theta_\mathcal{M}, \theta_\pi} \mathcal{L}$.

At prediction time, all we do is use the model consisting of the parameters $V^*, \theta_\mathcal{M}^*, \theta_\pi^*$. That is, we swap out the English embeddings of the full task model for French ones; this is depicted on the right-hand side of Figure~\ref{fig:architecture}.

\subsection{Example Model Transfer: Sentiment Classification}\label{subsec:sentclass}

We now show how the general framework we have described can be used to easily transfer an English sentiment classifier to other languages. All we need is to define the model architecture $\mathcal{M}$ and the two loss functions $\mathcal{L}_{D_L}$ and $\mathcal{L}_{D_P}$.

But first a note on data granularity. In our exposition of Section~\ref{subec:frmwrk}, we referred to inputs as sequences of words. However, in sentiment classification we now consider reviews, which are sequences of sentences -- which are in turn, sequences of words. Because our labeled data is chunked into review-sized inputs, the parallel data only needs to be aligned at paragraph, rather than sentence or word level.

Our model architecture $\mathcal{M}$ is a hierarchical Recurrent Neural Network (RNN) that is depicted in Figure~\ref{fig:sentirnn}. Hierarchical RNN models for document-level classification have previously proven to be effective~\cite{yang2016hierarchical,zhou2016attention}. For reference, our architecture achieves an F1 score of 0.92 when trained and tested in English on our benchmark dataset (see Section~\ref{subsec:data}). The model consists of a Sentence RNN which converts sequences of words into sentence representations, and a Review-level RNN which converts a sequence of sentence representations into a task representation. The variant of RNN we use is Gated Recurrent Units (GRUs)~\cite{cho2014learning}.

Given the binary nature of the prediction task, the prediction layer is simply a sigmoid layer with one output neuron that computes the probability of a positive label: $\widehat{\mathcal{\pi}_i^+} = \sigma(\theta_{\pi}^T \cdot \mathcal{R}_{x_i}^\mathbb{T})$. The labeled loss is a cross-entropy loss:

\begin{equation}
\mathcal{L}_{D_L} = -\displaystyle \sum_{i = 1}^N \left(y_i \log\widehat{\mathcal{\pi}_i^+} + (1 - y_i) \log(1 - \widehat{\mathcal{\pi}_i^+}) \right)
\end{equation}

\noindent On the parallel side, the unlabeled loss is a mean-squared error loss:

\begin{equation}
\mathcal{L}_{D_P} = \displaystyle \sum_{j = 1}^M \frac{1}{d^\mathbb{T}} \sum_{i = 1}^{d^\mathbb{T}} \left( \mathcal{R}_{e_j}^\mathbb{T}(i) - \mathcal{R}_{f_j}^\mathbb{T}(i) \right)^2
\end{equation}

\noindent where $d^\mathbb{T}$ is the dimension of the task representation $\mathcal{R}^\mathbb{T}$, and $\mathcal{R}^\mathbb{T}(i)$ denotes its $i^\text{th}$ dimension.
\section{Evaluation}\label{sec:eval}

We evaluate the transfer of the sentiment classifier we proposed in Section~\ref{subsec:sentclass} to three different languages on a publicly available dataset. We also show how the model may be combined with a simple translation baseline to yield state-of-the-art results. We begin by providing implementation details and describing the data.

\subsection{Implementation Details}\label{subsec:impl}

We implement our framework in Keras\footnote{\url{https://keras.io/}} and TensorFlow\footnote{\url{https://www.tensorflow.org/}}. The code will be made available by December 2018 at \url{https://github.com/Microsoft/CLModelTransfer}.

In our standard model, we set the embedding size to 64, and the encoding size of both sentence-level and review-level GRUs to 256. We use padding and truncation to normalize all input data to a uniform 30 sentences, each with 20 words. We set the dropout factor to 0.5~\cite{srivastava2014dropout} for all but the embedding layer. In two-stage training, we train the sentiment classifier on English labeled data for 10 epochs, then freeze its parameters, followed by 10 epochs of training in the parallel data. In the joint training setting, we pre-train the sentiment classifier on English labeled data for 4 epochs, followed by 12 epochs of joint training over labeled and parallel data. The optimizer we use is RMSProp~\cite{tieleman2012lecture}. We additionally experimented with varying several of the model hyperparameters\footnote{We can report that the following did not affect model performance, or only affected it adversely: using bidirectional instead of unidirectional GRUs, adding Gaussian Noise between layers, varying the mixing factor $\alpha$, and greatly changing the number of training epochs.}. A study of model size is presented in Section~\ref{subsec:size}.

\subsection{Data}\label{subsec:data}

We evaluate on the multilingual review dataset from~\cite{prettenhofer2010cross}, which consists of Amazon product reviews for books, DVDs and music in 4 different languages: English, German, French and Japanese. Reviews labeled $>3$ stars are considered positive, while those with $<3$ stars are tagged as negative. The reviews across languages are not aligned in any way.

We use the test portions of the data (consisting of a balanced set of 6000 reviews in each of the three languages) to measure precision, recall, F1 and accuracy in German, French and Japanese. On the English side, we use all available non-test instances from the dataset as labeled training data. This amounts to a total of 111,220 reviews. We apply the logic described above to convert ``star-ratings'' into binary positive or negative labels.

We use the Europarl corpus~\cite{koehn2005europarl} for English-German and English-French language pairs to obtain unlabeled parallel data. For each of these, the corpus contains approximately 2 million sentence pairs. In the English-Japanese language pair we use a parallel subtitle corpus~\cite{pryzant_jesc_2017}, which consists of 3.2 million sentence pairs. The subtitle corpus consists of much shorter elements, which are sometimes text fragments or even individual words.

While all the parallel data is aligned at \emph{sentence} level, our model does not actually need such fine-grained alignments. Thus we mimic paragraph-level alignment by batching contiguous groups of sentences into pseudo paragraphs. We vary the group size randomly between 15 and 30, and ignore internal alignment structure. As described in Section~\ref{subsec:impl}, padding and truncation are used to normalize the size of parallel pseudo paragraphs into consistent inputs.

\subsection{Baselines}\label{subsec:blines}

We compare against a number of baselines, which we divide into three broad categories.

\begin{table*}
    \centering
    \smallsize
    \begin{tabular}{c|cc|c|cccc|cccc|cccc}
    \hline 
        &  &  &  & \multicolumn{4}{c|}{DE} & \multicolumn{4}{c|}{FR} & \multicolumn{4}{c}{JP}\tabularnewline
        & \multicolumn{2}{c|}{Model} & \multicolumn{1}{c|}{Var} & Prec & Rec & F1 & Acc (\%) & Prec & Rec & F1 & Acc (\%) & Prec & Rec & F1 & Acc (\%)\tabularnewline
    \hline 
        & \multirow{4}{*}{TFTrans} & \multirow{2}{*}{LR} & TrnT & 0.670 & 0.906 & 0.770 & 73.0 & 0.736 & 0.847 & 0.788 & 77.2 & 0.574 & 0.957 & 0.718 & 62.4\tabularnewline
        &  &  & TstT & 0.759 & 0.774 & 0.766 & 76.4 & 0.779 & 0.704 & 0.739 & 75.2 & 0.572 & 0.888 & 0.695 & 61.1\tabularnewline
        &  & \multirow{2}{*}{NN} & TrnT & 0.727 & 0.852 & 0.785 & 76.6 & 0.774 & 0.747 & 0.760 & 76.5 & 0.604 & 0.797 & 0.688 & 63.8\tabularnewline
    Trans- &  &  & TstT & 0.814 & 0.651 & 0.723 & 75.1 & 0.819 & 0.624 & 0.708 & 74.3 & 0.640 & 0.516 & 0.572 & 61.3\tabularnewline
    lation & \multirow{4}{*}{MSTrans} & \multirow{2}{*}{LR} & TrnT & 0.810 & 0.840 & 0.825 & 82.2 & 0.836 & 0.794 & 0.814 & 81.9 & 0.637 & 0.93 & 0.756 & 70.0\tabularnewline
        &  &  & TstT & 0.804 & 0.879 & 0.840 & 83.2 & 0.873 & 0.770 & 0.818 & 82.9 & 0.766 & 0.784 & \textbf{0.775} & 77.2\tabularnewline
        &  & \multirow{2}{*}{NN} & TrnT & 0.870 & 0.854 & 0.862 & 86.3 & 0.909 & 0.728 & 0.809 & 82.8 & 0.787 & 0.754 & 0.770 & 77.5\tabularnewline
        &  &  & TstT & 0.909 & 0.830 & \textbf{0.868*} & \textbf{87.4*} & 0.920 & 0.759 & \textbf{0.832*} & \textbf{84.6*} & 0.857 & 0.669 & 0.751 & \textbf{77.8*}\tabularnewline
    \hline 
    No & \multicolumn{2}{c|}{Bi-SG} & - & 0.595 & 0.450 & 0.512 & 57.2 & 0.685 & 0.229 & 0.344 & 56.2 & 0.507 & 0.726 & \textbf{0.597} & 51.1\tabularnewline
    Trans- & \multicolumn{2}{c|}{Label} & 2-Stg & 0.583 & 0.784 & 0.669 & 61.1 & 0.595 & 0.884 & 0.711 & 64.1 & 0.807 & 0.386 & 0.522 & 64.7\tabularnewline
    lation & \multicolumn{2}{c|}{Projection} & Joint & 0.567 & 0.961 & \textbf{0.713} & \textbf{61.4} & 0.625 & 0.924 & \textbf{0.746} & \textbf{68.5} & 0.820 & 0.462 & 0.591 & \textbf{68.0}\tabularnewline
    \hline 
        & \multicolumn{2}{c|}{CL-SCL} &  & \multicolumn{3}{c}{-} & 78.0 & \multicolumn{3}{c}{-} & 78.4 & \multicolumn{3}{c}{-} & 73.1\tabularnewline
    %    & \multicolumn{2}{c|}{BSE} &  & \multicolumn{3}{c}{-} & 78.5 & \multicolumn{3}{c}{-} & - & \multicolumn{3}{c}{-} & 74.3\tabularnewline
    Prior & \multicolumn{2}{c|}{CR-RL} &  & \multicolumn{3}{c}{-} & 78.1 & \multicolumn{3}{c}{-} & 77.3 & \multicolumn{3}{c}{-} & 72.9\tabularnewline
    Work & \multicolumn{2}{c|}{Bi-PV} &  & \multicolumn{3}{c}{-} & 80.2 & \multicolumn{3}{c}{-} & 81.3 & \multicolumn{3}{c}{-} & 74.2\tabularnewline
        & \multicolumn{2}{c|}{Bi-DRL} &  & \multicolumn{3}{c}{-} & \textbf{84.3} & \multicolumn{3}{c}{-} & \textbf{83.5} & \multicolumn{3}{c}{-} & \textbf{76.2}\tabularnewline
    \hline 
    This  & \multicolumn{2}{c|}{Repr.} & 2-Stg & 0.708 & 0.912 & 0.797 & 76.8 & 0.703 & 0.918 & 0.796 & 76.6 & 0.773 & 0.711 & 0.741 & 75.2\tabularnewline
    Paper & \multicolumn{2}{c|}{Projection} & Joint & 0.796 & 0.816 & \textbf{0.806} & \textbf{80.4} & 0.782 & 0.842 & \textbf{0.811} & \textbf{80.4} & 0.735 & 0.846 & \textbf{0.786*} & \textbf{77.0}\tabularnewline
    \hline
    %Native & \multicolumn{2}{c|}{Upper Bound} &  & 0.940 & 0.904 & \textbf{0.911} & \textbf{92.3} & 0.947 & 0.907 & \textbf{0.927} & \textbf{92.9} & 0.913 & 0.880 & \textbf{0.896} & \textbf{89.8}\tabularnewline
    %\hline
    \end{tabular}
    \caption{Results on the Amazon Multilingual Reviews Dataset. Models are divided into 4 groups, and the best results from each group are highlighted in bold. Overall best results are additionally marked with *.}\label{tab:bigtable}
\end{table*}

\subsubsection{Translation Baselines}\label{subsubsec:trblines}

We build models over two translation systems. The first is a commercial solution from Microsoft (MSTrans). This system is trained on orders of magnitude more data than is available to us, and is highly tuned. So, in order to make a fairer comparison, we also train a state-of-the-art Neural Machine Translation (NMT) architecture on the same parallel data available to us (TFTrans). We use an encoder-decoder architecture, which consists of 3 layers of bidirectional LSTMs in both encoder and decoder; the embedding size is set to 128, while the dimensions of the hidden representations in the encoders are set to 512.

We distinguish 2 variants of translations. Training-time Translation (TrnT), which translates the English training data into another language and then trains a sentiment model in that language; and Test-time Translation (TstT), which trains an English sentiment model and uses it to classify reviews that are translated into English at test time.

Finally we also experiment with 2 modeling approaches. The first is a neural model (NN) based on the architecture in Figure~\ref{fig:sentirnn}. The second is a logistic regressor with L2 regularization that is trained on uni-, bi- and tri-gram features.

This yields a total of 8 translation baselines: the cross-product of 2 types each of translation, training/testing variants, and modeling approaches.

\subsubsection{No Translation Baselines}\label{subsubsec:notrblines}

We also build 3 baselines that do not rely on translation. The first leverages bilingual embeddings. Specifically we train a bilingual skip-gram model (Bi-SG) on our parallel data sets based on~\cite{luong2015bilingual}, and use the resulting English side embeddings as fixed features in training the English sentiment model.

The second and third models are based on Label Projection. This is similar to our architecture for model transfer, but instead of using task representations as soft supervision for the unlabeled loss, we use the predicted label on the English side of the parallel data as hard supervision. We distinguish 2 variants of this approach: 2-stage training, and joint training (see Section~\ref{subsec:training}).

\begin{table*}
    \centering
    \smallsize
    \begin{tabular}{c|cccc|cccc|cccc}
    \hline 
        & \multicolumn{4}{c|}{DE} & \multicolumn{4}{c|}{FR} & \multicolumn{4}{c}{JP}\tabularnewline
        & Prec & Rec & F1 & Acc (\%) & Prec & Rec & F1 & Acc (\%) & Prec & Rec & F1 & Acc (\%)\tabularnewline
    \hline 
    MSTranslator (TrnT) & 0.870 & 0.854 & 0.862 & 86.3 & 0.909 & 0.728 & 0.809 & 82.8 & 0.787 & 0.754 & 0.770 & 77.5\tabularnewline
    + Repr. Projection (Joint) & 0.873 & 0.860 & 0.866 & 86.7 & 0.906 & 0.772 & \textbf{0.834} & \textbf{84.6} & 0.790 & 0.803 & \textbf{0.797} & \textbf{79.5}\tabularnewline
    \hline 
	MSTranslator (TstT) & 0.909 & 0.830 & 0.868 & 87.4 & 0.920 & 0.759 & 0.832 & 84.6 & 0.857 & 0.669 & 0.751 & 77.8\tabularnewline
	+ Repr. Projection (Joint) & 0.903 & 0.847 & 0.874 & 87.8 & 0.909 & 0.792 & \textbf{0.846} & \textbf{85.6} & 0.852 & 0.697 & \textbf{0.767} & \textbf{78.8}\tabularnewline
    \hline 
    \end{tabular}
    \caption{Results of combining our approach with the MSTranslator NN model. Statistically significant improvements are shown in bold.}\label{tab:combo}
\end{table*}

\subsubsection{Prior Work Baselines}\label{subsubsec:prwrk}

Finally, we also compare against prior work. Here we cite~\cite{zhou2016cross}, who have the current state-of-the-art on the Multilingual Amazon Reviews dataset. All previous work on this dataset only reports accuracy scores, so we are only able to cite these numbers. \cite{zhou2016cross} report results from a number of different models that we also provide as reference points. Notably, all these approaches require some form of machine translation - either at training time, test time, or both.

\noindent \paragraph{CL-SCL:} The cross-lingual structural correspondence learning approach by~\cite{prettenhofer2010cross} that set the original benchmark results for the Multilingual Reviews dataset.

\noindent \paragraph{CR-RL:} A bilingual embedding method from~\cite{xiao2014semi} that shares knowledge across languages by having vectors that are part language-dependent, part language-independent.

\noindent \paragraph{Bi-PV:} An application of~\cite{pham2015learning}, who propose learning bilingual paragraph vectors by treating paragraph encodings as context.

\noindent \paragraph{Bi-DRL:} The bilingual document representation learning model by~\cite{zhou2016cross}, that is the current state-of-the-art. It leverages monolingual and bilingual constraints in order to jointly train a classifier, but requires both training and test data to be translated.

\subsection{Results}\label{subsec:results}

The results of our evaluation are presented in Table~\ref{tab:bigtable}. There are several interesting findings here. First, the NN model based on MSTrans produces very strong results. In fact, both TrnT and TstT variants of this simple baseline outperform the previous state-of-the-art (Bi-DRL) across all 3 languages. Note that Bi-DRL also uses a commercial translation engine (Google) at both training and test time. These improvements can be attributed to the progress that MT has made;~\cite{prettenhofer2010cross} report results from an MT baseline using Google translate, and the results from 2010 are significantly worse than effectively the same baseline in our work.

However, high quality translation comes at a cost. Translating our English training set into \emph{one} other language costs approximately \$1000, based on Microsoft Translator pricing; costs are similar for Google's translation service. Translation at test time is even worse in the long run, since costs are at least amortized in TrnT models. Costs can easily sky-rocket, if translation of large datasets into many languages is considered.

Meanwhile, using a state-of-the-art NMT architecture trained on publicly available data (TFTrans) produces significantly worse results. Because of the discrepancies between training and test data, introduced by poorer translations, we notice that LR can sometimes outperform NN models. This is in contrast to MSTrans, where higher translation quality means that NN yields better results. The performance of TrnT and TstT are also reversed vis-\`a-vis MSTrans and TFTrans; while TrnT performs slightly better in the case of TFTrans, TstT performs better for MSTrans.

The No Translation baselines are worse still. Label projection uses hard supervision that affects model performance dramatically. This is likely because the binary labels we use as weak supervision in parallel data do not necessarily \emph{mean} anything, and therefore cause the model to learn parameters that cannot generalize to actually meaningful test reviews. Meanwhile, Bi-SG is little better than random. We hypothesize that cross-lingual embeddings -- if trained in isolation -- are not able to mesh with higher layers of neural networks, which are often fragilely co-ordinated with one another~\cite{yosinski2014transferable}.

Meanwhile, our approach -- especially under joint training -- performs significantly better. We outperform the No Translation and TFTrans baselines by large margins. The differences are statistically significant, measured by Fisher's Exact Test ($\text{p-value} = 0.01$). We also outperform all prior work, except the state-of-the-art Bi-DRL model. In fact, in Japanese we even outperform Bi-DRL and rival MSTrans-NN on accuracy, and produce the best F1 score. This shows the robustness of our approach, which is less susceptible to language divergence than translation-based models.

In summary, our simple yet flexible framework produces results that are significantly better than other resource reliant, complex models, which require translation or curated dictionaries. We even rival state-of-the-art approaches that rely on high quality commercial translation engines, which have recourse to orders of magnitude more data.

\subsection{Model Composition with Translation}\label{subsec:compos}

While we are not able to outperform the MSTrans baseline, we examine if we can improve on it by combining our approach with translation. We test this by training a simple interpolation between our joint model and the MSTrans-NN models on training data. We then evaluate on the test sets. The results are presented in Table~\ref{tab:combo}.

This reveals that we are able to improve upon 2 of the 3 evaluation languages. The differences on German -- while higher numerically -- are not statistically significant. The results for French and Japanese are state-of-the-art. Thus our approach, while capable even without a translation engine, can also be used in combination with a commercial translator to yield state-of-the-art results.

As a reference, the empirical upper-bound yielded by native training and testing -- that is, using train and test data of the same language from the dataset -- is about 92\% for German and French, and 90\% for Japanese. An interesting avenue for future research is to explore better model composition techniques that could potentially close the gap with the upper-bound.
\section{Model Analysis}\label{sec:analysis}

Next we analyze our models both quantitatively and qualitatively.

\subsection{Effects of Model Size}\label{subsec:size}

\begin{figure}
\centering
\includegraphics[width=0.49\textwidth]{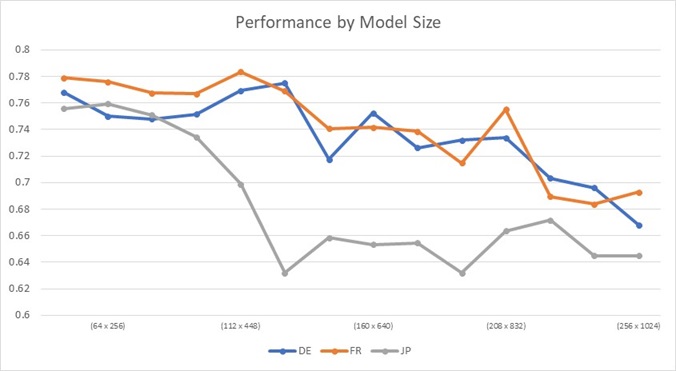}
\caption{Performance by model size. Results are the average \% accuracy over 10 runs.}
\label{fig:sizegraph}
\end{figure}

We first consider the effects of model size on cross-lingual transfer. We vary the embedding dimensions from 48 to 256 in increments of 16, and encoding dimensions from 256 to 1024 in increments of 64. For each size setting we run a total of 10 experiments and take the average accuracy across runs. The results of this experiment are presented in Figure~\ref{fig:sizegraph}.

They show that there is a general downward trend in accuracy with larger models. This is likely due to two factors. The first is that we have limited labeled training data, and with more parameters, overfitting is always an issue. Second, the unlabeled loss (Equation~\ref{eq:parloss}) is effectively using a vector (noisily generated by the sentiment model on unlabeled data) as weak supervision. The larger this vector, the more \emph{noise} it may contain, which in turn leads to less reliable supervision.

Thus our framework for model transfer is better suited to more compact models, unless bigger training datasets are available.

\subsection{Cross-Lingual Word Association}\label{subsec:viz}

We finally consider the embeddings learned via representation projection. Table~\ref{tab:viz} shows a few sentiment bearing English words, and their nearest French neighbors (by vector cosine distance on their respective embeddings) in our joint model.

\begin{table}
\centering
\small
\begin{tabular}{c|c}
\hline 
EN Word & Nearest FR Neighbors\tabularnewline
\hline
\multirow{2}{*}{excellent} & honneur, essentiel, heureux, formidable, fantastique,\tabularnewline
    & m'am\`ene, ravie, remarquable, bienvenue, heureuse\tabularnewline
\hline 
\multirow{2}{*}{terrible} & ridicule, gaspillage, d\'esol\'e, d\'echets, d\'eception,\tabularnewline
    & d\'ecevant, d\'e\c{c}u, honte, d\'esol\'ee, pire\tabularnewline
\hline 
\multirow{2}{*}{useful} & complet, fonctionnera, civils, appropri\'ee, utile,\tabularnewline
    & digne, utilit\'e, pertinent, pertinents, requises\tabularnewline
\hline 
\multirow{2}{*}{cheat} & d\'esole, mensonger, d\'echirer, insultant, erron\'es, \tabularnewline
    & d\'eficitaire, floue, comprendrais, impr\'egn\'e, incoh\'erent\tabularnewline
\hline 
\end{tabular}
\caption{The 10 closest French words (by embedding distance) to sentiment bearing English words.}\label{tab:viz}
\end{table}

We can see that positive (or negative) sentiment terms in English are close to positive (or negative) terms in French. Note that nearest neighbors are not necessarily accurate translations -- the sentiment prediction task does not really require translations; it is sufficient to identify words that echo the same sentiment. Our approach is thus able to pick up on sentiment similarity across languages, without direct supervision and only using the weak fuzzy signal from representation projection.
\section{Conclusion and Future Work}\label{sec:conclusion}

In this paper, we have shown a general framework for transferring neural models from one language to another. The framework leverages the novel idea of representation projection: using intermediate task representations as a form of weak predictive supervision. Notably our task agnostic framework requires little by way of additional resources (a parallel corpus is sufficient) and engineering (the only requirement is a loss function over task representations in parallel data). We show how this framework can be used to easily adapt an existing English neural model for sentiment classification to other languages. We test our approach on three target languages and evaluate on a public dataset of product reviews. The results reveal that our model is able to outperform a number of strong baselines, and rival more complex models that have access to significantly more training resources. When combined with a commercial translation-based model, we produce state-of-the-art results. Finally, our approach is able to learn embeddings that transfer the semantics of sentiment polarity across languages, without the need of direct supervision.

There are several avenues of future work that we plan to tackle. Since what we have proposed in this paper is a framework rather than a single model, we hope to investigate it's application to other problems and domains. Specifically we hope to tackle NLP tasks at different levels of granularity, such as sentence or word level classification problems. Our goal is to also extend the framework in order to accommodate more complex structured prediction problems, such as part-of-speech tagging and dependency parsing, as well as tackle multiple problems at once in a multi-task manner. Finally, we're also excited by the possibility of applying the framework to multi-modal (instead of multi-lingual) transfer, where -- for example -- one might transfer sentiment understanding to images, without the need for explicit image annotations.

\bibliographystyle{aaai}
\bibliography{biblio}

\end{document}